\documentclass{article} %
\usepackage{iclr2027_conference_arxiv,times}

\usepackage{amsmath,amsfonts,bm}

\def\eqref#1{equation~\ref{#1}}

\def\1{\bm{1}}

\DeclareMathAlphabet{\mathsfit}{\encodingdefault}{\sfdefault}{m}{sl}
\SetMathAlphabet{\mathsfit}{bold}{\encodingdefault}{\sfdefault}{bx}{n}

\usepackage[hidelinks]{hyperref}
\usepackage{url}
\usepackage{microtype}
\usepackage{booktabs}
\usepackage{fourier} 
\usepackage{array}
\usepackage[export]{adjustbox}
\usepackage{import}
\usepackage{float}
\usepackage{amsmath}
\usepackage{siunitx}
\usepackage{framed}
\usepackage{enumitem}
\usepackage{caption}
\usepackage{footnote}
\usepackage{footmisc}
\usepackage{lineno}
\usepackage{amssymb}
\usepackage{pifont}
\usepackage{multirow}
\usepackage{graphicx}
\DeclareMathAlphabet{\mathcal}{OMS}{cmsy}{m}{n}

\newcommand{\highlight}[1]{\setlength{\fboxsep}{1pt}\colorbox{yellow!80}{#1}}
\newcommand{\highlightgreen}[1]{\setlength{\fboxsep}{0pt}\colorbox{green!40}{#1}}

\definecolor{mygray}{gray}{0.4}

\title{Sliding-window beats linear attention}

\author{Alexia Jolicoeur-Martineau \\
Microsoft \\
Applied Sciences Group (ASG) \\
\texttt{alexiaj@microsoft.com} \\
\And
Rhea Sanjay Sukthanker \\
Microsoft \\
Applied Sciences Group (ASG) \\
\texttt{rsukthanker@microsoft.com} \\
\AND
Pashmina Cameron \\
Microsoft \\
Applied Sciences Group (ASG) \\
\texttt{pcameron@microsoft.com} \\
\And
\hspace{72pt} Emy Gervais \\
\hspace{72pt} Independent \\
\hspace{72pt} \texttt{emy.gervais0@gmail.com} \\
}

\iclrfinalcopy %
\fancypagestyle{plain}{\fancyhf{}\fancyfoot[C]{\thepage}\renewcommand{\headrulewidth}{0pt}}

\begin{document}

\maketitle

\begin{abstract}

Due to the nature of quadratic attention, Large Language Models (LLMs) consume a lot of memory and energy: every new token costs more than the previous one, and its keys and values must be stored in memory indefinitely, which is unsustainable. Two main lines of work address this: compressing the KV cache, e.g., by evicting or quantizing keys and values, and retrofitting LLMs to use Linear Attention, which replaces the KV cache with a fixed-size state. Retrofitting has attracted a lot of attention, given its promise to solve the quadratic scaling problem with state-of-the-art performance at low cost. However, it has not been properly compared to the simplest form of KV-cache eviction: Sliding Window Attention (SWA) with attention sinks. In this work, we show that SWA with sinks performs as well or better than most retrofitted Linear Attention models across multiple LLMs and downstream tasks, with the largest gains on long-context and generative tasks. On long-context reasoning tasks (Needle-in-a-Haystack and BABILong), SWA achieves massively higher performance (2 to 10 times higher than linear attention). SWA requires no additional training, is extremely fast, and requires little memory, making it an extremely cheap and reliable solution. When the training budget is limited, switching to SWA is a much more effective way to reduce inference memory cost than retrofitting linear attention. Linear attention models have shown promise, but they require training from scratch or extensive retrofitting to reap their architectural benefits and come close to SWA.

\end{abstract}

\begin{table}[ht]
    \centering
    \begin{adjustbox}{max width=\linewidth}
        \begin{tabular}{lcccc} 
 \toprule
 & Retrofitting & Retrofitting & \multicolumn{2}{c}{Recovery (\%)} \\
Name & Tokens & Stages & MMLU$\uparrow$ & Average$\uparrow$   \\
\midrule
SUPRA \citep{mercat2024linearizinglargelanguagemodels} & 100B & 1 & 53.0 (0.0) & 88.1 (0.0) \\
Hedgehog \citep{zhang2024hedgehogporcupineexpressive} & 40M & 2 & 36.9 (0.0) & 73.9 (0.0)  \\
LoLCATs \citep{zhang2024lolcatslowranklinearizinglarge} & 40M & 2 & 83.2 (2.2) & 97.5 (1.3)  \\ %
Liger-GLA \citep{lan2025liger} & 20M & 1 & 62.2 (5.8) & 92.0 (2.8)  \\
MOHAWK \citep{bick2024transformersssmsdistillingquadratic} & 3-5B & 3 & 56.9 (0.0) & 92.4 (0.0)  \\ %
Mamba in the Llama \citep{junxiongdaniele2024mambainllama} & 20B & 2 & 67.7 (0.0) & 86.7 (0.0) \\
DiJiang$^b$ \citep{Chen2024DiJiangEL} & 40B & 1 & 88.7 (0.0) & -  \\
ARWKV \citep{yueyu2025arwkvpretrainneedrnnattentionbased} & 60M/830M & 2/3 & 84.1 (0.0) & 94.7 (0.0) \\
Llamba \citep{bick2025llambascalingdistilledrecurrent} & 8-12B & 3 & 91.5 (0.0) & 98.6 (0.0) \\
QLinAtt \citep{goldstein2025radlads} & 350-700M & 3 & 74.0 (0.0) & 92.9 (0.0)  \\
QRWKV6 \citep{goldstein2025radlads} & 350-700M & 3 & 92.4 (2.7) & \highlightgreen{99.1} (0.8)  \\
QRWKV7 \citep{goldstein2025radlads} & 350-700M & 3 & 86.4 (6.8) & 96.1 (4.1)  \\
\highlight{Sliding Window Attention (64 window-size, 4 sinks)} & \highlightgreen{0} & \highlightgreen{0} & \highlightgreen{93.2} (3.5) & 99.0 (0.5)  \\
\bottomrule
\end{tabular}
\end{adjustbox}
\caption{\centering{Performance recovery (student score divided by teacher score) on MMLU-5shot and the average of 6 benchmarks (MMLU-5shot, ARC-C, ARC-E, Hellaswag, PIQA, Winogrande). \protect\linebreak}}
\label{tab:comparing_recovery}
\end{table}

\clearpage

\section{Introduction}

Large Language Models (LLMs) consume massive amounts of memory and energy because they scale quadratically with context-length \citep{vaswani2017attention}. Every new token’s keys and values must be added to the KV cache, and they contribute to an ever-growing memory and compute cost.

A large body of work reduces this cost by compressing the KV cache, for example by evicting less important tokens \citep{zhang2023h2o, liu2023scissorhands, li2024snapkv} or by loading only the most relevant entries at each step \citep{tang2024quest}. The simplest such policy keeps only the first few tokens (attention sinks) and a sliding window of recent tokens \citep{xiao2024efficient}, which is exactly the SWA with sinks baseline that we study.

Linear attention methods \citep{katharopoulos2020lineartransformrers} propose an alternative which reduces the time and memory complexity from quadratic to linear. This removes the need to store a KV cache which makes the memory cost small and fixed at inference-time instead of having an ever-growing KV cache. 

While promising, linear attention models come with significant drawbacks: 1) lower expressivity, 2) intractable problem of having to decide what to remember so that important information is not overwritten/ignored, and 3) training is expensive and most software/hardware are not built for them. 

To tackle the expensive training, a solution is to take an existing pretrained model and swap its expensive quadratic attention with a linear attention. This would normally require fine-tuning with billions of tokens \citep{bick2024transformers}, but LoLCATs \citep{zhang2024lolcatslowranklinearizinglarge} showed that this could be done with as little as 40M tokens by combining Hedgehog \citep{zhang2024hedgehogporcupineexpressive} and Sliding Window Attention (SWA). In doing so, LoLCATs recovers most of the full attention performance on knowledge and reasoning tasks.

Given the promises of these linearizing methods, we gave them a try. What we found was striking. We discovered that changing the attention mask to Sliding Window Attention (SWA) with attention sinks (meaning that each token attends to the first 4 tokens and the $w-4$ most recent tokens, for a window of $w$ tokens) gives us better downstream performance than most linearized models. For long-context tasks, the performance gap is even more pronounced, whereas SWA obtains much higher accuracy than linear attention retrofitting.

The fact that SWA with sinks retains most of the performance of the original model is well known \citep{xiao2024efficient, cabannes2026short}, however, to our knowledge, linearizing methods have not been compared to SWA with sinks. In this work, we provide this direct comparison for a wide variety of models (ranging from 1.3B to 72B) on various benchmarks (short and long context). In doing so, we demonstrate that pretrained models can already use SWA at inference time to obtain high performance at a fixed inference memory cost without needing any post-training or specialized linear kernels. 

\textbf{Notation} We denote Sliding Window Attention (SWA) in the following way: \text{SWA($w$, $s$)}, where $w$ is the window-size (varies from 64 to 512) and $s$ is the number of attention sinks (always fixed at 4).

\begin{figure}[htbp]
  \centering
  \includegraphics[width=\textwidth]{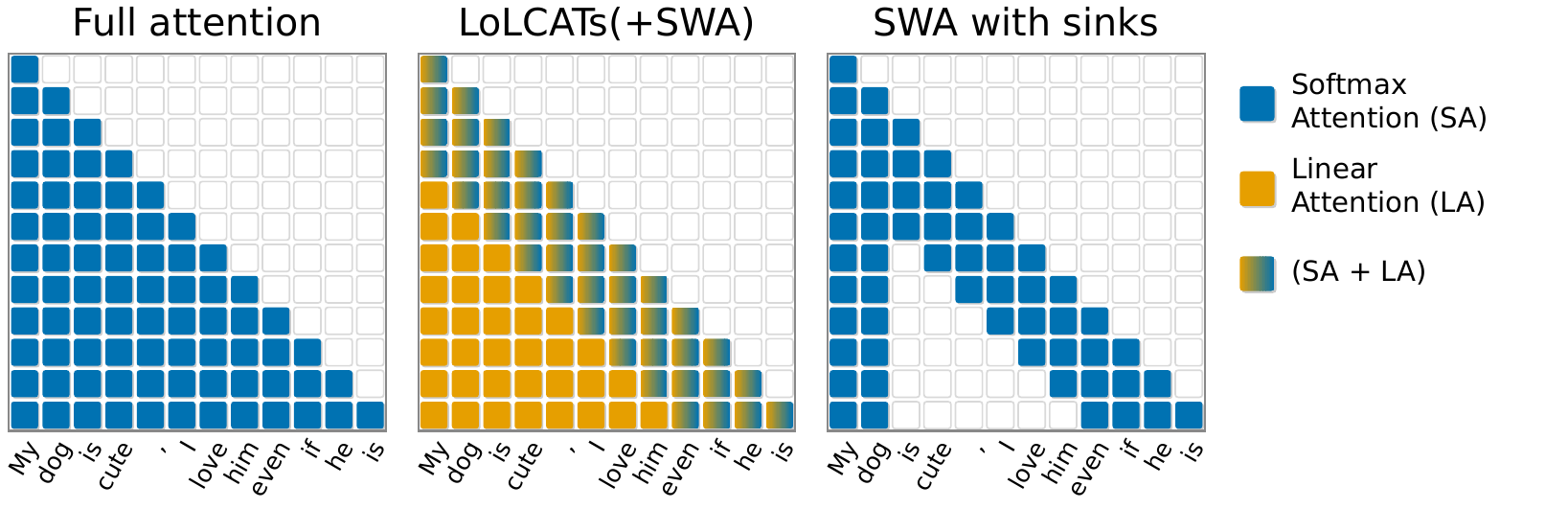}
  \caption{Different types of attention masks: Full attention (FA), LoLCATs/Liger-GLA (mixture of linear attention and SWA), and Sliding Window Attention (SWA) with sinks (adapted from Figure~1 of \citet{meng2026still}).}
  \label{fig:att_types}
\end{figure}

\section{Background}

\subsection{Transformers and Self-Attention (SA)}

Transformers \citep{vaswani2017attention} are the backbone of modern Large Language Models (LLMs). They process a sequence of $L$ consecutive tokens. Assuming discrete tokens (with vocab-size $V$) used in language, the tokens are embedded ($ V \to D$), transforming the shape of the sequence to $[L,D]$. Transformers then process this sequence by alternating between a transformation on the dimension $D$ using Multi-Layer Perceptron (MLP) and on the dimension $L$ using Self-Attention (SA). Both MLP and SA use residual connections ($x \gets x + f(x)$ where $f$ is either the MLP or SA). The final output is then projected back into shape $[L,V]$ to predict the next-token probabilities.

Self-Attention works as follows: the data of shape $[L,D]$ is linearly projected to keys $\mathbf{k}$, queries $\mathbf{q}$, and values $\mathbf{v}$ and the output is transformed in the following way:
\begin{equation}
    \mathbf{x}_{t} = \frac{\sum_{i=1}^t a_{t, i}\mathbf{v}_i}{\sum_{i=1}^t a_{t, i}} = \frac{\sum_{i=1}^t \exp(\mathbf{q}_t \mathbf{k}_i^\top / \sqrt{d})\mathbf{v}_i}{\sum_{i=1}^t \exp(\mathbf{q}_t \mathbf{k}_i^\top / \sqrt{d} )}, \;\;\;
    \text{for $t$ in $[1, \ldots, L]$}
\label{eq:full_attention}
\end{equation}

Several variants of the formulation have been proposed, such as Multi-Query Attention \citep{shazeer2019fast} and Grouped Query Attention \citep{ainslie2023gqa}. Unlike MLP which has complexity $\mathcal{O}(LD^2)$, Self-Attention has complexity $\mathcal{O}(DL^2)$ and it requires a KV cache at inference time to keep track of all previous keys and values in order to make the inference cost per new token linear $\mathcal{O}(L)$. The quadratic scaling with respect to context-length and the ever-growing KV cache is a big problem for LLMs as it increases in memory and requires more processing time as the sequence grows. 

Many engineering tricks have been exploited to make self-attention more efficient \citep{dao2022flashattention, dao2023flashattention, kwon2023efficient}, and a large body of work compresses the KV cache, e.g., through token eviction \citep{zhang2023h2o, liu2023scissorhands, li2024snapkv} or query-aware sparsity \citep{tang2024quest}. SWA with attention sinks (Section~\ref{sec:swa}) can be seen as the simplest KV-cache eviction policy. However, the quadratic problem remains. Various solutions have been suggested, but we focus on the following two: Sliding Window Attention (SWA) and Linear Attention (LA).

\subsection{Sliding Window Attention (SWA)}\label{sec:swa}

Instead of attending to all previous tokens, Sliding Window Attention (SWA) \citep{beltagy2020longformer} proposes to only attend to the previous $w$ tokens (its window size). This may appear extremely constraining, but similar to convolutional networks \citep{luo2016understanding}, the effective receptive field grows after each self-attention layer such that after $l$ layers, the receptive field is $lw$. Empirically, SWA improves long-term memorization and length extrapolation by encouraging models to learn dependencies beyond their local receptive field \citep{cabannes2026short}. SWA can be represented as follows:
\[
\mathbf{x}_t
=
\frac{\displaystyle\sum_{i=\max(1,t-w+1)}^{t}
\exp\!\left(\mathbf{q}_t\mathbf{k}_i^\top/\sqrt{d}\right)\mathbf{v}_i}
{\displaystyle\sum_{i=\max(1,t-w+1)}^{t}
\exp\!\left(\mathbf{q}_t\mathbf{k}_i^\top/\sqrt{d}\right)},
\qquad t\in[1,\ldots,L].
\]

It was later discovered that LLMs assign disproportionately high attention to the first few tokens even though they are not semantically relevant \citep{xiao2024efficient,barbero2025llms}. These tokens, called the \emph{attention sinks}, are tokens that transformers learn to use as repositories for any unnecessary attention. Without their inclusion in the attention mask, there is a catastrophic degradation in performance. Thus, when SWA moves past the first few (sink) tokens, performance becomes horrendous. A simple fix was later found: attend to the first $s=4$ tokens in addition to the remaining $w-4$ sliding window \citep{xiao2024efficient}. This approach solves the catastrophic failure that arises after the first tokens are out of the sliding window. 

SWA naturally works without any training, but additional post-training can boost its performance \citep{yu2025swaa}. There also exists learnable attention sinks \citep{agarwal2025gpt}, but these require additional post-training. In this work, we focus exclusively on training-free SWA with sinks. 

\subsection{Linear attention}

Linear Attention \citep{katharopoulos2020lineartransformrers} has been proposed as a way to alleviate the issues in Self-Attention. The main idea is to design a transformation $\phi$ such that $\exp(\mathbf{q}_t \mathbf{k}_i^\top) \approx \phi(\mathbf{q}_t) \phi(\mathbf{k}_i)^\top$. Then, the computations can be reformulated as: 
\begin{equation}
    \mathbf{x}_{t} = \frac{\sum_{i=1}^t \phi(\mathbf{q}_t) \phi(\mathbf{k}_i)^\top\mathbf{v}_i}{\sum_{i=1}^t \phi(\mathbf{q}_t) \phi(\mathbf{k}_i)^\top } = \frac{\phi(\mathbf{q}_t)\sum_{i=1}^t  \phi(\mathbf{k}_i)^\top\mathbf{v}_i}{\phi(\mathbf{q}_t)\sum_{i=1}^t  \phi(\mathbf{k}_i)^\top } = \frac{\phi(\mathbf{q}_t) \mathbf{s}_t}{\phi(\mathbf{q}_t) \mathbf{z}_t },
\label{eq:linear_attention1}
\end{equation}

Thus at inference, the equation becomes:
\begin{equation}
    \mathbf{s}_t=\mathbf{s}_{t-1}+\phi(\mathbf{k}_t)^\top\mathbf{v}_t,\quad\mathbf{z}_t=\mathbf{z}_{t-1}+\phi(\mathbf{k}_t)^\top.
    \label{eq:linear_attention2}
\end{equation}
This makes the inference cost $\mathcal{O}(1)$ with respect to $L$ since it does not depend on the sequence length. Thus one only needs to store and update $\mathbf{s}$ and $\mathbf{z}$ over time, both of which do not grow in size over time.

This resolves the issue of increasing memory and lower speed over long context lengths. However, making Linear Attention work well in practice is extremely challenging due to the nature of having to continuously rewrite itself while trying not to forget/ignore important information.

Many variants of linear attention with varying performance have been proposed \citep{gu2023mamba, qin2022cosformer, sun2023retentive, peng2023rwkv, de2024griffin, yang2024parallelizing, dao2024transformers,yang2024gatedlinearattentiontransformers, rwkv6_colm, peng2025rwkv}. Furthermore, multiple kernels have been proposed such as ReLU or ELU \citep{katharopoulos2020lineartransformrers}. There are 3 necessary properties that are required for a good kernel: expressiveness, spikiness, and monotonicity. To obtain these properties, Hedgehog \citep{zhang2024hedgehogporcupineexpressive} proposed using a learnable projection followed by a dual-sided exponential transformation: 
\begin{equation}
\phi(x) \gets (\exp(f(x)), \exp(-f(x)))
    \label{eq:hedgehog}
\end{equation}
where $f$ is a linear projection from $D$ to $D/2$.

\subsubsection{Retrofitting linear attention}

Training linear attention models is expensive and most software/hardware is made for Softmax attention, which makes things more challenging. Instead of training from scratch a linear attention Transformer, one can convert existing pretrained LLMs with quadratic Self-Attention to use linear-attention instead. With some additional training, which we refer to as \emph{retrofitting}, a large portion of the baseline’s performance can be recovered. Most importantly, LoLCATs \citep{zhang2024lolcatslowranklinearizinglarge} showed that one could linearize models with as little as 40M tokens using Low-Rank Adaptation (LoRA) \citep{hu2021lora}. To make it work, they use an expressive kernel such as Hedgehog \citep{zhang2024hedgehogporcupineexpressive} and combine the linear attention with a small Sliding Window Attention (SWA). This idea paved the way toward extremely efficient methods for retrofitting linear attention models.

\section{The missing comparison: Sliding Window vs Linear Attention}

Retrofitting linear attention is an interesting way of solving the quadratic scaling problem, although it comes with its own drawbacks (e.g., low expressivity, intractable problem of knowing what to retain and forget, requires additional training). However, most linear attention papers only compare linear-attention to regular (sink-free) SWA. This makes the comparison inadequate since we know that sink-free SWA leads to catastrophic failure once the first (sink) tokens are out of the sliding window. Furthermore, the long-context behavior of retrofitted linear attention models is understudied.

In this work, we demonstrate that Sliding Window Attention (SWA) with attention sinks performs comparably to or better than most retrofitted linearized models on short-context knowledge and reasoning tasks. SWA leads to especially large gains on long-context. See our results below.

\section{Experiments}

We compare existing linearizing methods (pure linear or linear with SWA) to the base pretrained model and their sliding-window attention (SWA) analog. For simplicity and fairness, we do not compare to hybrid models containing some full-attention layers. %

\subsection{General Knowledge and Reasoning}

We compare the different linear attention methods listed in Table \ref{tab:comparing_recovery} to Sliding Window Attention (SWA; window-size=64 with 4 sinks) on general knowledge and reasoning metrics with different base models (See Appendix \ref{app:details} for details). The summarized metrics are shown in Table \ref{tab:comparing_recovery} and the full results in Table \ref{tab:short_context}.

\begin{table}[ht]
    \centering
    \begin{adjustbox}{max width=\linewidth}
        \begin{tabular}{lrr|cccccc|c} 
\toprule
 &  & Retrofitting  & MMLU$\uparrow$ & ARC-C$\uparrow$ & ARC-E$\uparrow$ & HellaSwag$\uparrow$ & PIQA$\uparrow$ & WinoGrande$\uparrow$ & Avg $\uparrow$ \\
Model & Type & Tokens (B) & (5-shot) & (\text{acc-norm}) & (acc) & (\text{acc-norm}) & (acc) & (acc) &  \\
\midrule
\href{https://huggingface.co/microsoft/phi-1_5}{Phi-1.5-1.3B}& Teacher & 0 & 42.5 & 48.0 & 76.2 & 62.6 & 76.6 & 72.9 & 63.1 \\
& SWA(64, 4) & 0 & \highlightgreen{39.3} & \highlightgreen{48.1} & 76.2 & 61.5 & 76.4 & \highlightgreen{72.8} & 62.4 \\
& MOHAWK & 3 &24.2&44.1&74.0&60.2&75.5&71.7 & 58.3\\
& \highlightgreen{LoLCATs(+SWA)} & 0.04 &39.2&46.9&\highlightgreen{77.0}&\highlightgreen{62.3}&\highlightgreen{76.9}&72.7  &  \highlightgreen{62.5}\\
\midrule
\href{https://huggingface.co/mistralai/Mistral-7B-v0.1}{Mistral-7B-v0.1} & Teacher & 0 & 62.5 & 54.3 & 80.1 & 81.2 & 80.8 & 75.1 & 72.3 \\
& \highlightgreen{SWA(64, 4)} & 0  & \highlightgreen{56.3} & 53.8 & 80.3 & 80.6 & 80.9 & \highlightgreen{75.4} & \highlightgreen{71.2} \\
& SUPRA & 20 &	33.1 & 45.8 & 75.9 & 77.1 & 80.1 & 70.3 & 63.7 \\
& LoLCATs(+SWA) & 0.04 &51.4&\highlightgreen{54.9}&\highlightgreen{81.7}&\highlightgreen{80.7}&\highlightgreen{81.5}&74.0 & 70.7 \\
& Liger-GLA(+SWA) & 0.02 &36.3&49.3&78.7&76.3&80.1&70.1 & 65.1 \\
\midrule
\href{https://huggingface.co/meta-llama/Llama-2-7b}{Llama-2.0-7B} & Teacher & 0 & 45.9 & 45.1 & 75.5 & 76.2 & 78.1 & 69.3 & 65.0 \\ 
& \highlightgreen{SWA(64, 4)} & 0 & 39.8 & \highlightgreen{44.8} & \highlightgreen{75.5} & \highlightgreen{75.3} & \highlightgreen{78.1} & \highlightgreen{69.5} & \highlightgreen{63.8} \\
& DiJiang & 40 &\highlightgreen{40.7}&42.7&62.6&69.4&77.5&- & - \\
\midrule
\href{https://huggingface.co/meta-llama/Meta-Llama-3-8B}{Llama-3.0-8B} & Teacher & 0 & 65.5 & 53.9 & 80.8 & 79.1 & 78.5 & 73.3 & 71.9 \\
& \highlightgreen{SWA(64, 4)} & 0 & \highlightgreen{59.8} & 54.2 & 80.9 & 78.8 & 78.8 & 73.7 & \highlightgreen{71.0} \\
& Hedgehog & 0.04 &24.2&40.6&71.1&50.7&77.4&54.3 & 53.1 \\
& LoLCATs(+SWA) & 0.04 &52.8&\highlightgreen{54.9}&\highlightgreen{81.7}&\highlightgreen{79.7}&\highlightgreen{80.9}&\highlightgreen{74.1} & 70.7 \\
& Liger-GLA(+SWA) & 0.02 &43.4&52.5&81.1&76.3&80.3&72.0 & 67.6 \\
\midrule
\href{https://huggingface.co/meta-llama/Meta-Llama-3-8B-Instruct}{Llama-3.0-8B-Instruct}  \hspace{-30pt} & Teacher & 0  & 66.8 & 55.9 & 82.2 & 75.5 & 77.7 & 71.3 & 71.6 \\
& \highlightgreen{SWA(64, 4)} & 0  & \highlightgreen{61.4} & \highlightgreen{55.7} & \highlightgreen{82.2} & \highlightgreen{75.3} & \highlightgreen{78.1} & \highlightgreen{71.3} & \highlightgreen{70.7} \\
& Mamba2(+SWA) & 0 &45.2&48.0&74.1&70.8&75.8&58.6 & 62.1 \\
\midrule
\href{https://huggingface.co/meta-llama/Llama-3.1-8B}{Llama-3.1-8B} & Teacher & 0  & 65.4 & 55.1 & 81.9 & 79.2 & 78.9 & 74.2 & 72.5 \\
& \highlightgreen{SWA(64, 4)} & 0   & \highlightgreen{60.6} & \highlightgreen{55.1} & 82.1 & 78.9 & 79.3 & \highlightgreen{75.0}  & \highlightgreen{71.8} \\
& LoLCATs(+SWA) & 0.04  & 54.9&54.4&82.4&\highlightgreen{79.1}&\highlightgreen{81.0}&69.7 & 70.3 \\
 & Llamba &  12  & 60.0&54.6&\highlightgreen{82.5}&77.6&80.9&73.3 & 71.5 \\
\midrule
\href{https://huggingface.co/meta-llama/Llama-3.1-70B}{Llama-3.1-70B} & Teacher & 0 & 78.9 & 61.4 &85.0&85.7&82.5&81.6 & 79.1 \\
& \highlightgreen{SWA(64, 4)} & 0 & \highlightgreen{73.2} & \highlightgreen{61.2} &\highlightgreen{85.1}&\highlightgreen{85.4}&\highlightgreen{82.3}&\highlightgreen{81.7} & \highlightgreen{78.2} \\
& LoLCATs(+SWA) & 0.04 &67.7&60.5&85.0&84.6&82.1&73.7 & 75.6 \\
\midrule
\href{https://huggingface.co/Qwen/Qwen2.5-7B-Instruct}{Qwen2.5-7B-Instruct} \hspace{-30pt} & Teacher & 0  & 74.2 & 55.4 & 81.4 & 80.4 & 79.7 & 70.9 & 73.7 \\
& \highlightgreen{SWA(64, 4)} & 0  & \highlightgreen{70.3} & 55.4 & \highlightgreen{81.8} & \highlightgreen{79.8} & 79.4 & 71.3 & \highlightgreen{73.0} \\
& ARWKV & 0.02 &62.4&52.2&79.7&76.8&79.2&68.7 & 69.8 \\
& QLinAtt & 0.6 &54.9&53.4&79.6&75.3&78.8&68.9 & 68.5 \\
& QRWKV6-RoPE & 0.6 &66.1&\highlightgreen{56.7}&81.7&78.9&79.9&70.6 & 72.3 \\
& QRWKV7 & 0.6 &65.7& 56.3& 81.4& 79.0& \highlightgreen{80.3}& 71.1 & 72.3 \\
& QRWKV7-RoPE & 0.6 &68.2&55.8&81.7&79.3&79.8&\highlightgreen{71.8}  & 72.8 \\
\midrule
\href{https://huggingface.co/Qwen/Qwen2.5-32B-Instruct}{Qwen2.5-32B-Instruct} \hspace{-30pt} & Teacher & 0 & 83.2 & 58.5 & 82.2 & 85.3 & 80.8 & 73.0 & 77.2 \\
& SWA(64, 4) & 0 & \highlightgreen{79.5} & 58.8 & 81.9 & \highlightgreen{84.9} & 80.7 & 73.6 & 76.6 \\
& \highlightgreen{QRWKV6} & 0.6 &76.6&\highlightgreen{60.9}&\highlightgreen{84.3}&83.0&\highlightgreen{81.2}&\highlightgreen{78.2} & \highlightgreen{77.3} \\
\midrule
\href{https://huggingface.co/Qwen/QwQ-32B}{QwQ-32B} & Teacher & 0&79.9&55.6&81.0&84.1&79.8&70.5 & 75.2 \\
& \highlightgreen{SWA(64, 4)} & 0 &\highlightgreen{79.9}&55.6&\highlightgreen{81.0}&\highlightgreen{84.1}&79.8&70.5 & \highlightgreen{75.2} \\
& QRWKV6 & 0.6 &74.3&\highlightgreen{56.4}&80.6&83.0&\highlightgreen{80.4}&\highlightgreen{73.2} & 74.7 \\
\midrule
\href{https://huggingface.co/Qwen/Qwen2.5-72B-Instruct}{Qwen2.5-72B-Instruct} \hspace{-30pt} & Teacher & 0 &84.6&63.1&86.0&87.4&83.3&76.4 & 80.1\\ 
& \highlightgreen{SWA(64, 4)} & 0 &\highlightgreen{81.8}&63.7&86.4&\highlightgreen{87.2}&\highlightgreen{83.2}&76.3 & \highlightgreen{79.8} \\
& QRWKV6 & 0.6 & 77.5& \highlightgreen{63.8}& \highlightgreen{86.5}& 85.7& 82.5& \highlightgreen{79.6} & 79.3 \\ 
& QRWKV7 & 0.6 &66.7&57.2&83.3&78.2&80.1&73.9  & 73.2 \\ 
\bottomrule
\end{tabular}
\end{adjustbox}
\caption{\centering{Benchmark scores (\%) for distilled linearized models, teachers, and teachers with SWA (4 sinks with sliding window 64). The best non-teacher model is \highlightgreen{highlighted}. \protect\linebreak
}}
\label{tab:short_context}
\end{table}

\textbf{Results:} SWA obtains the best average downstream performance in 9 out of 11 cases. The only exceptions are a) LoLCATs on Phi-1.5-1.3B which obtains a score barely higher (62.5 vs 62.4 for SWA), and b) QRWKV6 on Qwen2.5-32B-Instruct which performs as well as the baseline model (77.3) while SWA has a slight drop of performance (76.6). On MMLU, SWA is always the best performing except for Llama-2.0-7B where DiJiang is better (40.7 versus 39.8 for SWA). 

\textbf{Summarized results:} SWA recovers the most out of MMLU baseline's performance (93.2\%), followed closely by QRWKV6 (92.4\%). Both SWA and QRWKV6 recover most of the average baseline performance (99.0\% for SWA and 99.1\% for QRWKV6). In terms of training efficiency at high performance, SWA is the winner with 0 tokens required, followed by LoLCATs which fine-tunes on 40M tokens to recover 83.2\% of MMLU and 97.5\% of the average baseline performance.

As additional experiments, we also retrofit our own linearized models with several state-of-the-art attention variants on modern architectures. Similar results were found when comparing them to SWA; see Appendix \ref{app:swa-linear-comparison} for details.

\subsection{Generative Reasoning}

The benchmarks above are scored by log-likelihood. In Table \ref{tab:reasoning}, we evaluate generation on math (GSM8K \citep{cobbe2021gsm8k}) and code (HumanEval \citep{DBLP:journals/corr/abs-2107-03374}).

\begin{table}[ht]
\centering
\resizebox{0.9\textwidth}{!}{%
\begin{tabular}{ll|cc|c}
\toprule
 &  & GSM8K$\uparrow$ & HumanEval$\uparrow$ & Avg$\uparrow$ \\
Model & Type & (5-shot, EM) & (0-shot, pass@1) &  \\
\midrule
Qwen2.5-7B-Instruct & Teacher & 75.6 & 64.0 & 69.8 \\
 & SWA(64,4) & 0.0 & 12.8 & 6.4 \\
 & SWA(128,4) & 1.7 & 28.7 & 15.2 \\
 & SWA(256,4) & 25.4 & 40.2 & 32.8 \\
 & SWA(512,4) & \highlightgreen{41.7} & 39.6 & \highlightgreen{40.7} \\
 & QRWKV6 (linear) & 2.8 & 40.9 & 21.8 \\
 & QRWKV7 (linear) & 29.6 & \highlightgreen{45.1} & 37.4 \\
\midrule
Qwen2.5-7B & Teacher & 82.1 & 55.5 & 68.8 \\
 & SWA(64,4) & 0.5 & 9.8 & 5.2 \\
 & SWA(128,4) & 3.1 & 29.3 & 16.2 \\
 & SWA(256,4) & 33.6 & 32.9 & 33.3 \\
 & SWA(512,4) & \highlightgreen{46.6} & 33.5 & \highlightgreen{40.0} \\
 & QRWKV6-Base (linear) & 2.7 & \highlightgreen{38.4} & 20.6 \\
\midrule
Qwen2.5-32B-Instruct & Teacher & 75.1 & 54.9 & 65.0 \\
 & SWA(64,4) & 0.0 & 17.7 & 8.8 \\
 & SWA(128,4) & 4.8 & 40.2 & 22.5 \\
 & SWA(256,4) & 29.6 & 52.4 & 41.0 \\
 & SWA(512,4) & \highlightgreen{52.8} & \highlightgreen{54.3} & \highlightgreen{53.6} \\
 & QRWKV6 (linear) & 44.1 & 36.6 & 40.4 \\
\midrule
QwQ-32B & Teacher & 74.0 & 41.5 & 57.8 \\
 & SWA(64,4) & 0.1 & 15.9 & 8.0 \\
 & SWA(128,4) & 10.3 & 32.3 & 21.3 \\
 & SWA(256,4) & 45.9 & 40.9 & 43.4 \\
 & SWA(512,4) & \highlightgreen{58.8} & \highlightgreen{42.1} & \highlightgreen{50.5} \\
 & QRWKV (linear) & 17.2 & 30.5 & 23.9 \\
\midrule
Llama-3.1-8B & Teacher & 50.8 & 34.8 & 42.8 \\
 & SWA(64,4) & 2.7 & 11.0 & 6.8 \\
 & SWA(128,4) & 21.9 & 28.0 & 25.0 \\
 & SWA(256,4) & 41.9 & \highlightgreen{36.6} & 39.3 \\
 & SWA(512,4) & \highlightgreen{49.2} & 36.0 & \highlightgreen{42.6} \\
 & LoLCATs (linear) & 0.6 & 14.0 & 7.3 \\
\bottomrule
\end{tabular}}
\caption{Generative reasoning benchmarks (\%). Avg is the mean of GSM8K and HumanEval. The best non-teacher result per base model is \highlightgreen{highlighted}.}
\label{tab:reasoning}
\end{table}

\textbf{Results:} Generative tasks require larger windows: SWA at window-size 64 collapses on GSM8K, but performance improves steadily as the window-size increases. At window-size 512, SWA obtains the best average among train-free and linearized methods on every base model. Linearized models are competitive on HumanEval but mostly fail on GSM8K.

\subsection{Long Context Reasoning}

\subsubsection{Single Needle-in-a-Haystack (S-NIAH)}

We compare SWA, LoLCATs \citep{zhang2024lolcatslowranklinearizinglarge}, and Liger-GLA \citep{lan2025liger} on Single Needle-in-a-Haystack (S-NIAH) tasks at context lengths up to 4K. Baseline numbers are taken from \citet{meng2026still}. The base model used by all approaches is Llama-3.1-8B \citep{grattafiori2024llama}. We compare at different window-sizes (128, 256, 512). Results are shown in Table \ref{tab:niah}.

\begin{table}[ht]
    \centering
    \setlength{\tabcolsep}{10pt}
    \resizebox{1\linewidth}{!}{
        \begin{tabular}{l|c|cccc|cccc|cccc}
            \toprule
            (Base: Llama-3.1-8B)
            & Window
            & \multicolumn{4}{c|}{\textbf{\textsc{S-NIAH-1}}}
            & \multicolumn{4}{c|}{\textbf{\textsc{S-NIAH-2}}}
            & \multicolumn{4}{c}{\textbf{\textsc{S-NIAH-3}}}
            \\
            Model
            & size
            & .5K 
            & 1K 
            & 2K 
            & 4K 
            & .5K 
            & 1K 
            & 2K 
            & 4K 
            & .5K 
            & 1K 
            & 2K
            & 4K 
            \\
            \midrule 
            \highlightgreen{SWA(128,4)} & 128 & \highlightgreen{35.0} & \highlightgreen{20.2} & \highlightgreen{15.0} & \highlightgreen{12.6} & \highlightgreen{100} & \highlightgreen{33.0} & \highlightgreen{22.8} & \highlightgreen{9.2} & \highlightgreen{99.8} & \highlightgreen{56.2} & \highlightgreen{41.4} & \highlightgreen{17.2} \\
            LoLCATs(+SWA) & 128 & 29.4 & 9.6 & 3.0 & 0 & 100 & 17.4 & 7.2 & 4.2 & 98.2 & 14.6 & 3.2 & 1.6 \\
            Liger-GLA(+SWA) & 128 &  28.4 & 0.2 & 0.2 & 0.2 & 100 & 1.6 & 0.6 & 1.0 & 97.6 & 2.0 & 1.2 & 0.8 \\
            \midrule
            \highlightgreen{SWA(256,4)} & 256 & \highlightgreen{91.8} & \highlightgreen{34.8} & \highlightgreen{21.0} & \highlightgreen{14.6} & \highlightgreen{100} & \highlightgreen{51.4} & \highlightgreen{33.4} & \highlightgreen{13.8} & \highlightgreen{100} & \highlightgreen{68.0} & \highlightgreen{50.2} & \highlightgreen{19.6}  \\
            LoLCATs(+SWA) & 256 & 84.8 & 26.2 & 10.2 & 2.2 & 100 & 37.0 & 12.2 & 8.2 & 100 & 30.6 & 6.0 & 2.2 \\
            Liger-GLA(+SWA) & 256 & 83.8 & 0.0 &  0.0 & 2.8 & 100 & 1.0 & 4.6 & 0.8 & 97.6 & 1.0 & 3.8 & 0.6 \\
            \midrule
            \highlightgreen{SWA(512,4)} & 512 & \highlightgreen{100} & \highlightgreen{69.0} & \highlightgreen{33.6} & \highlightgreen{19.0} & \highlightgreen{100} & \highlightgreen{85.2} & \highlightgreen{51.0} & \highlightgreen{23.0} & \highlightgreen{100} & \highlightgreen{90.4} & \highlightgreen{64.8} & \highlightgreen{23.0} \\
            LoLCATs(+SWA) & 512 & 100 & 65.6 & 24.6 & 8.8 & 100 & 71.8 & 17.4 & 16.6 & 100 & 51.0 & 10.6 & 5.8 \\
            Liger-GLA(+SWA) & 512 & 73.4 & 1.0 & 0.0 & 0.0 & 100 & 2.6 & 4.0 & 0.0 & 97.6 & 1.2 & 1.2 & 0.0 \\
            \midrule
            Full Attention & $\infty$ & 100 & 100 & 100 & 100 & 100 & 100 & 100 & 100 & 100 & 99.8 & 100 & 99.8\\
            \bottomrule
        \end{tabular}
    }
    \caption{Accuracy on the \textbf{Single Needle-in-a-Haystack (S-NIAH)} across context lengths (0.5K, 1K, 2K, 4K) and window-size (128, 256, 512). The best model at each window-size is \highlightgreen{highlighted}.}
\label{tab:niah}
\end{table}

\textbf{Results:} At all window-sizes (128, 256, 512) and context-lengths, SWA obtains equal or higher scores on all tasks. On S-NIAH-3 at 4K context-length, SWA recovers 17.2--23\% of the full attention accuracy, while LoLCATs and Liger-GLA reach at most 5.8\% and 0.8\% accuracy respectively.

\subsubsection{BABILong}

We compare SWA and LoLCATs \citep{zhang2024lolcatslowranklinearizinglarge} on the BABILong benchmark \citep{kuratov2024babilong}, which hides the supporting facts of the 20 bAbI reasoning tasks \citep{weston2016babi} (e.g., ``Mary travelled to the office'') inside distractor text from PG-19 books \citep{rae2020compressive}, and asks a question about them (e.g., ``Where is Mary?''). Following \citet{meng2026still}, we report the average over the first five tasks: single supporting fact (QA1), two supporting facts (QA2), three supporting facts (QA3), two-argument relations (QA4), and three-argument relations (QA5). The context length is the total input length in tokens after adding distractor text; 0K denotes the original task without distractor text. We evaluate at context lengths up to 4K with window-size 256. Baseline numbers are taken from  \citet{meng2026still}. The base model is Llama-3.1-8B \citep{grattafiori2024llama}. Results are shown in Table \ref{tab:babilong2} (see Appendix \ref{app:babilong} for details).

\begin{table*}[ht]
    \centering
        \begin{tabular}{l|c|cccc}
            \toprule
            (Base: Llama-3.1-8B)
            & Window & \multicolumn{4}{c}{\textbf{\textsc{BABILong (Average)}}}
            \\ Model
            & size
            & 0K 
            & 1K 
            & 2K 
            & 4K 
            \\
            \midrule 
            SWA(256, 4) & 256 & 55 & 20 & \highlightgreen{19} & \highlightgreen{15}
            \\
            LoLCATs(+SWA) & 256 & \highlightgreen{56} & \highlightgreen{22} & 10 & 3 
            \\
            \midrule
            Full Attention & $\infty$ & 74 & 70 & 67 & 60 
            \\
            \bottomrule
        \end{tabular}
    \caption{Average accuracy (\%) on the first five \textbf{BABILong} tasks (QA1--QA5) across context lengths, from 0K (no distractor text) to 4K tokens. Base model: Llama-3.1-8B.}
    \label{tab:babilong2}
\end{table*}

\textbf{Results:} At small context tasks, LoLCATs(+SWA) obtains slightly higher scores than SWA (56\% versus 55\% at 0K and 22\% versus 20\% at 1K). At large context tasks, SWA obtains noticeably higher scores than LoLCATs (19\% versus 10\% at 2K and 15\% versus 3\% at 4K). At small context length (0K), SWA and LoLCATs(+SWA) recover 74-76\% of the baseline performance. At long context length (4K), SWA recovers 25\% of the baseline performance, unlike LoLCATs which recovers only 5\%.

\newpage
\subsubsection{LongBench, RULER, and BABILong at longer contexts}

Table \ref{tab:long-context} extends the long-context evaluation to LongBench \citep{bai2024longbench}, LongBench-v2 \citep{bai2024longbenchv2}, RULER \citep{hsieh2024ruler}, and BABILong (QA1--QA20) at up to 16K context length. Unlike Table~\ref{tab:babilong2}, which averages QA1--QA5 on Llama-3.1-8B at context lengths up to 4K with baseline numbers from \citet{meng2026still}, here we run all methods ourselves and average BABILong over all 20 tasks at 8K and 16K tokens, so the BABILong numbers in the two tables are not directly comparable.

\begin{table}[ht]
\centering
\resizebox{\linewidth}{!}{%
\begin{tabular}{ll|ccc|cc|c}
\toprule
 &  & LongBench$\uparrow$ & LongBench-v2$\uparrow$ & RULER$\uparrow$ & \multicolumn{2}{c|}{BABILong$\uparrow$} & Avg$\uparrow$ \\
Model & Type & (subtask avg) & (subtask avg) & (NIAH/VT/CWE/FWE) & (8K) & (16K) &  \\
\midrule
Qwen2.5-7B-Instruct & Teacher & 36.5 & 31.8 & 98.6 & 62.2 & 61.0 & 57.1 \\
 & SWA(64,4) & 12.8 & 22.0 & 4.0 & 31.5 & 32.7 & 17.7 \\
 & SWA(128,4) & 15.3 & 28.2 & 12.4 & 41.1 & 43.1 & 24.5 \\
 & SWA(256,4) & 14.7 & 28.3 & 17.7 & 41.9 & 43.3 & 25.8 \\
 & SWA(512,4) & \highlightgreen{17.2} & \highlightgreen{29.0} & \highlightgreen{22.7} & 43.6 & 44.2 & \highlightgreen{28.2} \\
 & QRWKV6 (linear) & 10.0 & 23.5 & 5.0 & 33.5 & 32.4 & 17.9 \\
 & QRWKV7 (linear) & 10.5 & 24.3 & 10.9 & \highlightgreen{48.4} & \highlightgreen{45.2} & 23.1 \\
\midrule
Qwen2.5-7B & Teacher & 32.3 & 30.9 & 99.2 & 43.2 & 41.1 & 51.1 \\
 & SWA(64,4) & 10.6 & 18.8 & 4.0 & 36.6 & \highlightgreen{38.6} & 17.8 \\
 & SWA(128,4) & 13.9 & 26.1 & 10.3 & 35.8 & 37.4 & 21.7 \\
 & SWA(256,4) & 15.1 & \highlightgreen{29.5} & 15.1 & 36.5 & 37.8 & 24.2 \\
 & SWA(512,4) & \highlightgreen{16.7} & 28.7 & \highlightgreen{17.8} & \highlightgreen{37.4} & 38.3 & \highlightgreen{25.3} \\
 & QRWKV6-Base (linear) & 10.2 & 27.0 & 4.0 & 28.7 & 27.1 & 17.3 \\
\midrule
Qwen2.5-32B-Instruct & Teacher & 37.9 & 36.9 & 99.9 & 72.6 & 71.6 & 61.7 \\
 & SWA(64,4) & 12.0 & 26.9 & 5.7 & 43.3 & 43.0 & 21.9 \\
 & SWA(128,4) & 13.2 & 28.5 & 13.2 & 52.0 & 51.3 & 26.6 \\
 & SWA(256,4) & 14.9 & \highlightgreen{35.0} & 19.8 & 53.1 & 51.8 & 30.5 \\
 & SWA(512,4) & 17.4 & 32.6 & \highlightgreen{26.7} & \highlightgreen{55.1} & \highlightgreen{52.8} & \highlightgreen{32.7} \\
 & QRWKV6 (linear) & \highlightgreen{27.3} & 28.8 & 19.3 & 39.4 & 38.4 & 28.6 \\
\midrule
QwQ-32B & Teacher & 36.0 & 33.2 & 98.2 & 15.5 & 15.1 & 45.7 \\
 & SWA(64,4) & 12.2 & 29.2 & 3.3 & \highlightgreen{8.4} & \highlightgreen{8.0} & 13.2 \\
 & SWA(128,4) & 14.0 & 33.7 & 11.0 & 5.7 & 5.2 & 16.0 \\
 & SWA(256,4) & 15.0 & \highlightgreen{35.5} & 16.5 & 6.5 & 5.6 & 18.3 \\
 & SWA(512,4) & \highlightgreen{17.5} & 34.4 & \highlightgreen{23.6} & 8.3 & 6.4 & \highlightgreen{20.7} \\
 & QRWKV (linear) & 16.8 & 31.3 & 4.3 & 3.0 & 2.0 & 13.7 \\
\midrule
Llama-3.1-8B & Teacher & 29.0 & 25.5 & 98.9 & 10.1 & 9.2 & 40.8 \\
 & SWA(64,4) & 11.8 & 21.3 & 2.2 & \highlightgreen{10.1} & \highlightgreen{9.7} & 11.3 \\
 & SWA(128,4) & 14.8 & 24.3 & 11.8 & 7.9 & 7.3 & 14.6 \\
 & SWA(256,4) & 16.8 & 24.4 & 18.6 & 7.5 & 6.9 & 16.8 \\
 & SWA(512,4) & \highlightgreen{18.7} & \highlightgreen{24.5} & \highlightgreen{24.5} & 7.7 & 7.0 & \highlightgreen{18.8} \\
 & LoLCATs (linear) & 6.0$^\dagger$ & 18.6$^\dagger$ & 0.9 & 0.2 & 0.1 & 6.4 \\
\bottomrule
\end{tabular}}
\caption{Long-context benchmarks (accuracy \%). BABILong is averaged over all 20 tasks (QA1--QA20); 8K and 16K denote the total input length in tokens. Avg is the mean of LongBench, LongBench-v2, RULER, and the BABILong average over context lengths. The best non-teacher result per base model is \highlightgreen{highlighted}. $^\dagger$Evaluated at 8K context (LoLCATs exceeds H100 memory at 16K).}
\label{tab:long-context}
\end{table}

\textbf{Results:} SWA(512,4) obtains the best average among fixed-memory methods on every base model, while LoLCATs fails almost completely. However, all fixed-memory methods remain far from full attention, especially on RULER.

\subsection{Speed and Memory}

We compare full attention (FA), Sliding Window Attention (SWA) with (4) attention sinks to linear attention and linear attention with SWA (LoLCATs) in terms of memory and speed. Results are shown in Figure \ref{fig:speed}. We also provide plots with additional details in Appendix \ref{app:plotsextra}. 

\begin{figure}[htbp]
  \centering
  \includegraphics[width=1.0\textwidth]{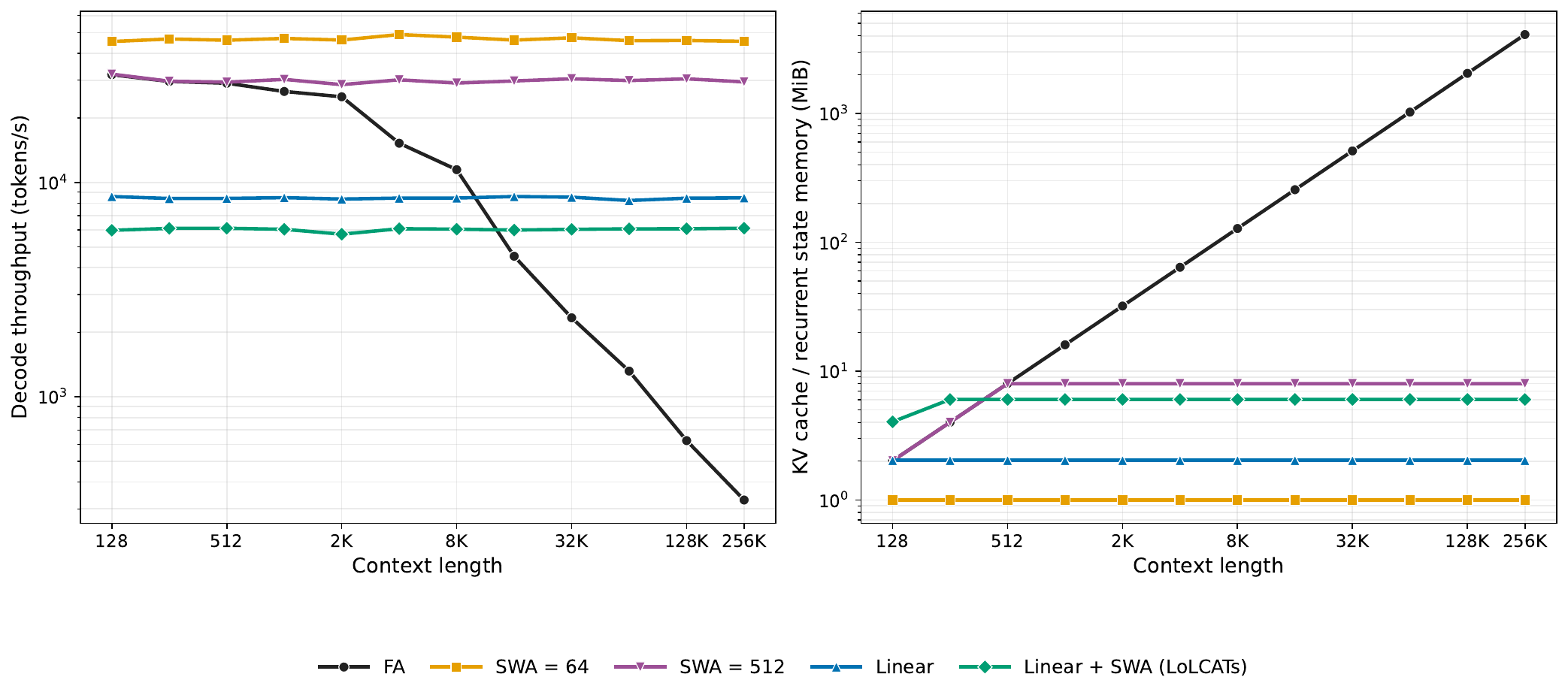}
  \caption{Speed (decoding throughput; tokens/s) and Memory cost (KV Cache or Recurrent state memory; MiB) of different attention types (Full Attention (FA), Sliding-Window Attention (SWA), Linear Attention, Linear + SWA (LoLCATs)) across context lengths (128 to 256K). \textbf{Both axes are in log scale.}}
  \label{fig:speed}
\end{figure}

\textbf{Hardware/Software:} FA and SWA both use FlashAttention \citep{dao2022flashattention,dao2023flashattention} as backend. For linear attention, ThunderKittens \citep{spector2024thunderkittens} is used as backend. For LoLCATs, the fused SWA + linear-attention kernel from ThunderKittens at window-size=256 is used. Tests are done on a 4-layer Transformer (embedding-size=1024, 16 heads with dimension 64) with batch-size=1 in float16 precision. The hardware is an NVIDIA RTX PRO 6000 Blackwell Max-Q Workstation Edition.

\textbf{Speed:} FA decreases in speed as context-length grows beyond 1K, since attention is quadratic in the sequence length by nature: each decoded token attends to all previous tokens, so its cost grows linearly with the context length (see Figure \ref{fig:flops}). All other methods stay with a relatively flat speed. SWA is the fastest method (window-size=64 is faster than 512, but both are faster than the other methods).

\textbf{Memory:} FA increases in memory cost linearly with context length. SWA increases in memory cost until it reaches its window-size, then it stays constant. SWA at window-size=64 has the lowest memory cost, followed by Linear, Linear + SWA (LoLCATs), and finally SWA at window-size=512.

Thus, SWA is faster and it has a similar or lower memory cost at window-size smaller than 512.

\section{Limitations and Future Work}

Our work is focused on training-free SWA. However, post-training can further improve the performance of SWA \citep{yu2025swaa}; future work should investigate the effect of post-training SWA and compare it to linear-attention retrofitting at different amounts of tokens (i.e., scaling laws). We also only consider the simplest KV-cache eviction policy (SWA with sinks). Score-based eviction methods such as H2O \citep{zhang2023h2o} or SnapKV \citep{li2024snapkv} can also run at a fixed memory budget and may perform even better than SWA; comparing them to retrofitted linear attention is an interesting direction for future work. We did not consider hybrid models with some tokens or entire layers using full attention (FA); it would be interesting to know the impact of different degrees of full attention with SWA or linear attention. The analysis could also be extended to extremely large models and complex metrics, such as agentic tasks. The effect on multi-modal LLMs and video generation models with a multi-dimensional sliding window (e.g., 3D with (x, y) image position and (t) time coordinates) would also be interesting to investigate. Note that for video diffusion models \citep{ho2022video}, there is already strong evidence that training-free SWA is an extremely strong baseline: for example, Sliding Tile Attention \citep{zhang2025fast} recovers 97\% of the baseline's VBench score \citep{huang2024vbench} on HunyuanVideo \citep{kong2024hunyuanvideo} at 3.53 times the speed with the same number of sampling steps.

\section{Conclusion}

In this work, we showed that Sliding Window Attention (SWA) with attention sinks achieves better performance than most linear attention retrofitting methods. It does so without any post-training, at higher decoding speed, and lower memory cost.

On short-context reasoning tasks, SWA recovers 99\% of the average baseline’s performance, which matches the more expensive state-of-the-art linear attention retrofitting methods. On long-context tasks, SWA performs overwhelmingly better than linear-attention retrofitting. With a window-size of 256 at 4K context length, SWA recovers 20\% and 25\% of baseline performance on S-NIAH-3 and BABILong while LoLCATs only recovers 2.2\% and 5\% respectively. On longer contexts (up to 16K) and on generative math and code tasks, SWA(512,4) obtains the best average among train-free and linearized methods for all base models tested, although generative tasks require a larger window than log-likelihood tasks.

Given our findings, we strongly recommend practitioners to use SWA with attention sinks in order to achieve the best performance at a fixed small memory cost.

\bibliography{paper}
\bibliographystyle{iclr2027_conference}

\appendix
\section{Appendix}

\subsection{List of models and metrics}\label{app:details}

We compare methods on the following pretrained architectures: Phi-1.5-1.3B \citep{textbooks2}, Mistral-7B-v0.1 \citep{jiang2023mistral7b}, Llama-2.0-7B \citep{touvron2023llama}, Llama-3.0-8B, Llama-3.0-8B-Instruct, Llama-3.1-8B, Llama-3.1-70B \citep{grattafiori2024llama}, Qwen2.5-7B, Qwen2.5-7B-Instruct, Qwen2.5-32B-Instruct, Qwen2.5-72B-Instruct  \citep{qwen2025qwen25technicalreport}, and QwQ \citep{qwq32b}. The size of the models ranges from 1.3B to 72B. The metrics considered are: MMLU \citep{hendrycks2021mmlu}, ARC-C, ARC-E \citep{clark2018thinkarc}, PIQA \citep{bisk2020piqa}, WinoGrande \citep{sakaguchi2021winogrande}, HellaSwag \citep{zellers2019hellaswagmachinereallyfinish}. We chose these metrics because they are the standards used by other papers, which makes comparisons easier \citep{zhang2024lolcatslowranklinearizinglarge, goldstein2025radlads}. For generative and long-context evaluation, we additionally use GSM8K \citep{cobbe2021gsm8k}, HumanEval \citep{DBLP:journals/corr/abs-2107-03374}, S-NIAH and RULER \citep{hsieh2024ruler}, LongBench \citep{bai2024longbench}, LongBench-v2 \citep{bai2024longbenchv2}, and BABILong \citep{kuratov2024babilong}. 

\subsection{Additional results on newly linearized models}\label{app:swa-linear-comparison}

Table~\ref{tab:swa-linear-comparison} compares SWA$(64,4)$ to linearized models that we retrofit ourselves with several state-of-the-art attention variants on modern architectures.

\begin{table*}[ht]
\centering
\resizebox{\textwidth}{!}{%
\begin{tabular}{lrr|cccccc|c}
\toprule
\textbf{Model} & \textbf{Type} &
\begin{tabular}[c]{@{}c@{}}\textbf{Retrofitting}\\\textbf{Tokens (B)}\end{tabular} &
\begin{tabular}[c]{@{}c@{}}\textbf{MMLU}\\\textbf{(5-shot)}\end{tabular} &
\begin{tabular}[c]{@{}c@{}}\textbf{ARC-C}\\\textbf{(acc-norm)}\end{tabular} &
\begin{tabular}[c]{@{}c@{}}\textbf{ARC-E}\\\textbf{(acc)}\end{tabular} &
\begin{tabular}[c]{@{}c@{}}\textbf{HellaSwag}\\\textbf{(acc-norm)}\end{tabular} &
\begin{tabular}[c]{@{}c@{}}\textbf{PIQA}\\\textbf{(acc)}\end{tabular} &
\begin{tabular}[c]{@{}c@{}}\textbf{WinoGrande}\\\textbf{(acc)}\end{tabular} &
\textbf{Avg.} \\
\midrule

\multirow{6}{*}{Qwen3-8B}
 & Teacher         & 0   & 74.9 & 56.7 & 83.5 & 75.0 & 76.6 & 68.4 & 72.5 \\
 & SWA$(64,4)$     & 0   & \highlightgreen{70.8} & \highlightgreen{56.9} &\highlightgreen{ 83.6} & \highlightgreen{74.3} & \highlightgreen{76.5} & \highlightgreen{67.6} & \highlightgreen{71.6} \\
 & Gated DeltaNet  & 0.1 & 27.5 & 46.8 & 76.6 & 59.1 & 73.9 & 52.9 & 56.1 \\
 & GLA             & 0.1 & 25.8 & 38.0 & 70.5 & 50.9 & 72.0 & 50.6 & 51.3 \\
 & QRWKV6          & 0.1 & 24.6 & 41.5 & 74.7 & 53.7 & 72.6 & 52.0 & 53.2 \\
\midrule

\multirow{6}{*}{Phi-4-mini-reasoning}
 & Teacher         & 0   & 57.4 & 48.0 & 71.4 & 64.8 & 69.3 & 59.1 & 61.7 \\
 & SWA$(64,4)$     & 0   & \highlightgreen{54.8} & \highlightgreen{48.1} & \highlightgreen{71.0} & \highlightgreen{63.3} & \highlightgreen{69.0} & \highlightgreen{59.5} & \highlightgreen{60.9} \\
 & Gated DeltaNet  & 0.1 & 24.3 & 34.7 & 64.0 & 45.5 & 68.7 & 51.1 & 48.1 \\
 & GLA             & 0.1 & 25.2 & 30.7 & 59.2 & 40.9 & 66.5 & 51.8 & 45.7 \\
 & QRWKV6          & 0.1 & 25.1 & 31.4 & 61.4 & 41.8 & 67.6 & 52.0 & 46.6 \\
\midrule

\multirow{6}{*}{Phi-4-reasoning-plus}
 & Teacher         & 0   & 77.8 & 56.5 & 82.5 & 78.7 & 80.1 & 76.7 & 75.4 \\
 & SWA$(64,4)$     & 0   & \highlightgreen{74.4} & \highlightgreen{56.5} & \highlightgreen{82.4} & \highlightgreen{78.4} & \highlightgreen{80.3} & \highlightgreen{77.1} & \highlightgreen{74.9} \\
 & Gated DeltaNet  & 0.1 & 25.9 & 44.5 & 75.3 & 60.1 & 75.7 & 52.6 & 55.7 \\
 & GLA             & 0.1 & 23.7 & 45.0 & 75.9 & 60.0 & 74.3 & 52.8 & 55.3 \\
 & QRWKV6          & 0.1 & 25.7 & 40.1 & 72.7 & 57.3 & 74.6 & 53.2 & 53.9 \\
\bottomrule
\end{tabular}%
}
\caption{Comparison of sliding-window and linear-attention methods. Teacher and SWA$(64,4)$ are train-free baselines; linear variants are LoLCATs-style two-stage distilled on $\sim$0.1B tokens of cleaned-Alpaca \citep{alpaca}. We use different attention variants (GLA \citep{yang2024gatedlinearattentiontransformers}, Gated DeltaNet \citep{yang2025gateddeltanetworksimproving}, QRWKV6 \citep{rwkv6_colm}) on modern architectures (Qwen3 \citep{yang2025qwen3}, Phi-4-mini-reasoning \citep{abdin2024phi, xu2025phi}, and  Phi-4-reasoning-plus \citep{abdin2025phi}).} 
\label{tab:swa-linear-comparison}
\end{table*}

\subsection{BABILong Full Results}\label{app:babilong}

Table~\ref{tab:babilong} reports the per-task BABILong results (QA1--QA5) summarized in Table~\ref{tab:babilong2}.

\begin{table*}[ht]
    \centering
    \resizebox{1\textwidth}{!}{
        \begin{tabular}{l|cccc|cccc|cccc|cccc|cccc}
            \toprule
            (Base: Llama-3.1-8B)
            & \multicolumn{4}{c|}{\textbf{\textsc{QA1}}}
            & \multicolumn{4}{c|}{\textbf{\textsc{QA2}}}
            & \multicolumn{4}{c|}{\textbf{\textsc{QA3}}}
            & \multicolumn{4}{c|}{\textbf{\textsc{QA4}}}
            & \multicolumn{4}{c}{\textbf{\textsc{QA5}}}
            \\ Model
            & 0K 
            & 1K 
            & 2K 
            & 4K 
            & 0K 
            & 1K 
            & 2K 
            & 4K 
            & 0K 
            & 1K 
            & 2K 
            & 4K 
            & 0K 
            & 1K 
            & 2K 
            & 4K 
            & 0K 
            & 1K 
            & 2K 
            & 4K 
            \\
            \midrule 
            SWA(256, 4) & 84 & 21 & \highlightgreen{16} & \highlightgreen{12} 
            & \highlightgreen{36} & \highlightgreen{14} & \highlightgreen{11} & \highlightgreen{6}
            & 26 & 16 & \highlightgreen{21} & \highlightgreen{11}
            & \highlightgreen{82} & 16 & \highlightgreen{17} & \highlightgreen{18}
            & 49 & 35 & \highlightgreen{29} & \highlightgreen{30}
            \\
            LoLCATs(256) & \highlightgreen{100} & \highlightgreen{22} & 5 & 3 
            & 25 & 10 & 4 & 1 
            & \highlightgreen{30} & \highlightgreen{17} & 13 & 4 
            & 65 & \highlightgreen{21} & 9 & 1 
            & \highlightgreen{62} & \highlightgreen{42} & 17 & 6 
            \\
            \midrule
            Full attention & 94 & 88 & 82 & 74 
            & 59 & 49 & 48 & 44
            & 47 & 53 & 48 & 44
            & 82 & 76 & 74 & 67
            & 90 & 86 & 83 & 73
            \\
            \bottomrule
        \end{tabular}
    }
    \caption{\textbf{BABILong Benchmark Full Results.}
    Accuracy (\%) on each of the first five BABILong tasks (QA1--QA5) across context lengths, from 0K (no distractor text) to 4K tokens. Base model: Llama-3.1-8B.
    }
    \label{tab:babilong}
\end{table*}

\clearpage

\subsection{Speed, Memory, and FLOPs plots}\label{app:plotsextra}

\begin{figure}[htbp]
  \centering
  \includegraphics[width=1.0\textwidth]{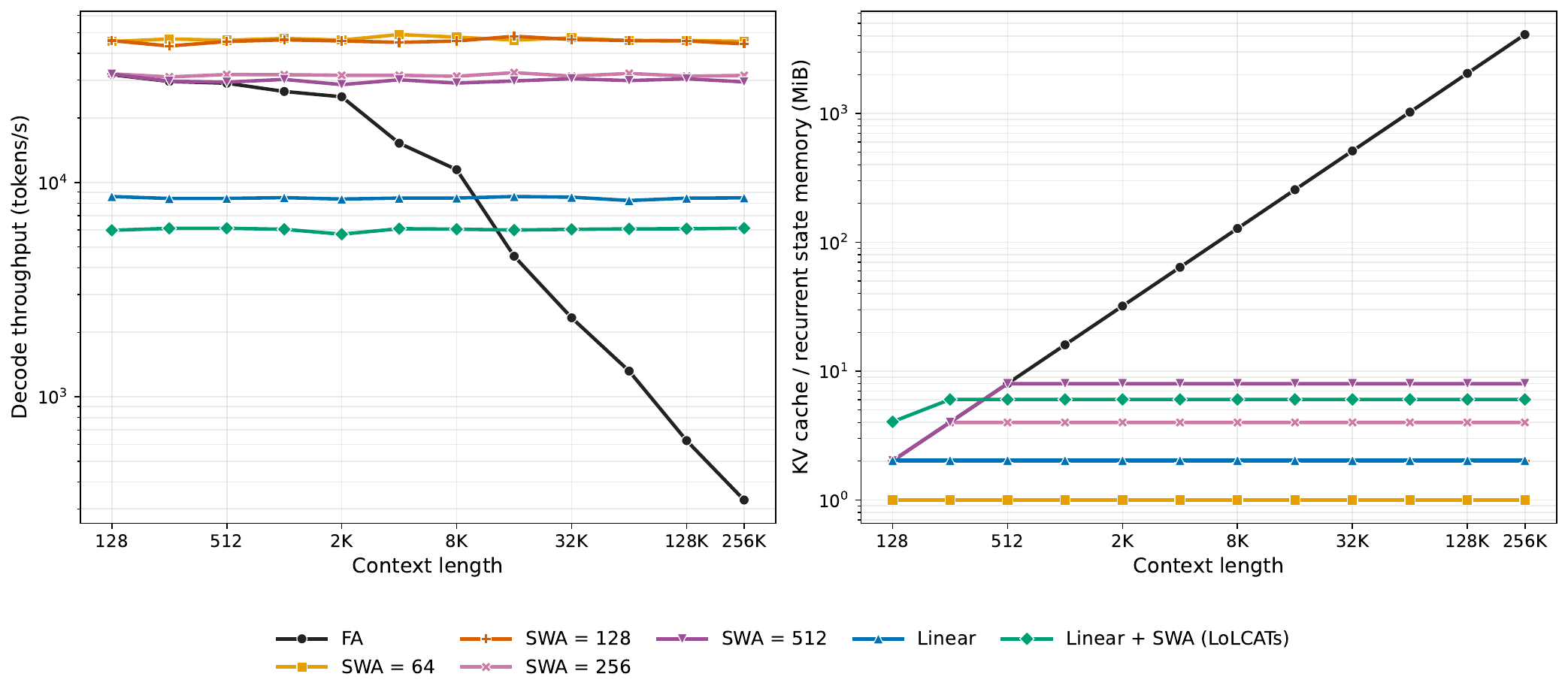}
  \caption{Speed (decoding throughput; tokens/s) and Memory cost (KV Cache or Recurrent state memory; MiB) of different attention types (Full Attention (FA), Sliding-Window Attention (SWA), Linear Attention, Linear + SWA (LoLCATs)) across context lengths (128 to 256K). \textbf{Both axes are in log scale.}}
  \label{fig:speed2}
\end{figure}

\begin{figure}[htbp]
  \centering
  \includegraphics[width=1.0\textwidth]{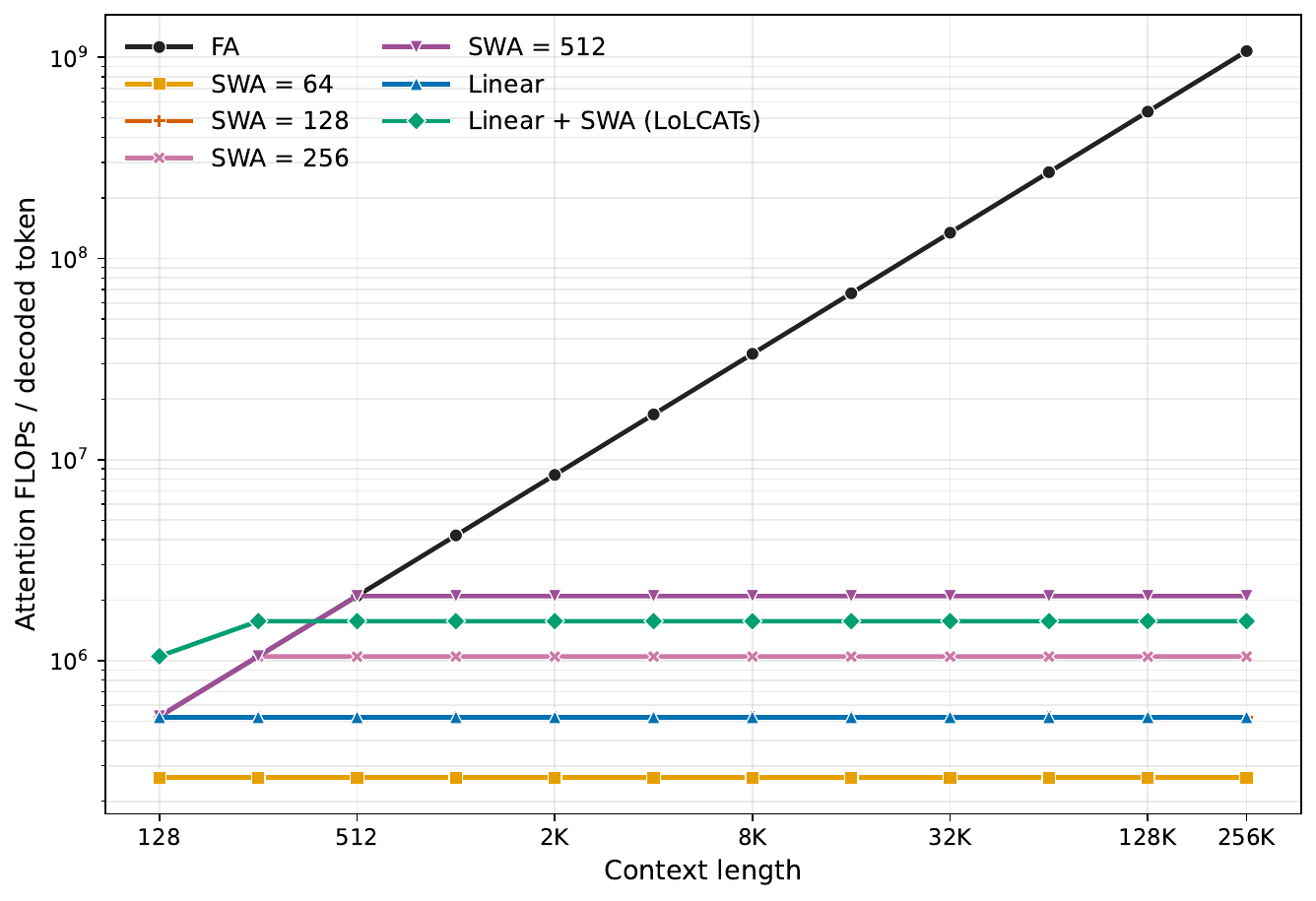}
  \caption{Attention floating point operations (FLOPs) per decoded token of different attention types (Full Attention (FA), Sliding-Window Attention (SWA), Linear Attention, Linear + SWA (LoLCATs)) across context lengths (128 to 256K). \textbf{Both axes are in log scale.}}
  \label{fig:flops}
\end{figure}

\end{document}